\documentclass{book}

\usepackage[utf8]{inputenc}
\usepackage{pdfpages}
\usepackage{fancyhdr}
\usepackage{hyperref}

\usepackage{combelow}
\usepackage{newunicodechar}

\usepackage{ amstext, epstopdf, booktabs, verbatim, gensymb, geometry, lmodern}
\usepackage{tocloft}
\newlength{\standardchapnumwidth}
\AtBeginDocument{\setlength{\standardchapnumwidth}{\cftchapnumwidth}}

\newunicodechar{Ș}{\cb{S}}

\newcommand*\cpiType{Volume 4}
\newcommand*\Date{December 2023}
\newcommand*\Author{Elfia Bezou-Vrakatseli \\ Federico Castagna \\ Isabelle Kuhlmann \\ Jack Mumford \\ \cb{S}tefan Sarkadi \\ Daphne Odekerken \\ Maddie Waller \\ Andreas Xydis \\}

\definecolor{myblue}{HTML}{154360}
\definecolor{emerald}{HTML}{3cb371}

\begin{document}

\newgeometry{margin = 0in}

\pagecolor{emerald}
\setlength{\fboxsep}{0pt}
\hfill \colorbox{myblue}{\makebox[3.22in][r]{\shortstack[r]{\vspace{3.3in}}}}%
\setlength{\fboxsep}{15pt}
\setlength{\fboxrule}{5pt}
\colorbox{white}{\makebox[\linewidth][c]{\includegraphics[width=1.3in]{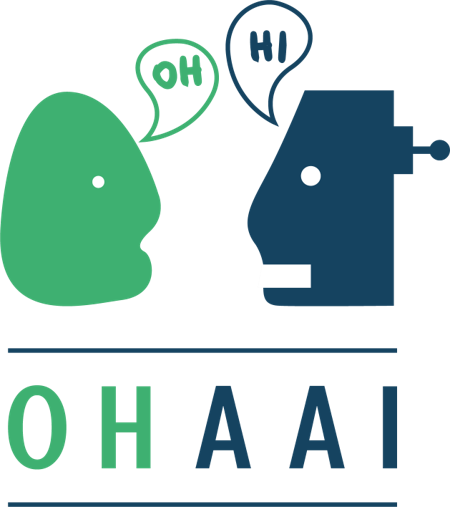}\hspace{0.35in} \shortstack[l]{\vspace{10pt}\fontsize{40}{40}\rmfamily\color{myblue} Online Handbook of \\
\vspace{10pt}
\fontsize{40}{40}\rmfamily\color{myblue} Argumentation for AI\\%
\fontsize{20}{20}\rmfamily\color{myblue} \cpiType}}}%
\setlength{\fboxsep}{0pt}
\vspace{-0.25pt}
\hfill \colorbox{myblue}{\hspace{.25in} \parbox{2.97in}{\vspace{1.4in} \color{white} \large{Edited by \\ \\ \Author  \\  \Date \vspace{2.15in} \vfill}}}%
\restoregeometry

\nopagecolor

\thispagestyle{empty}
\pagenumbering{gobble}

\begin{center}
    \textbf{{\huge Preface}}
\end{center}

\hfill

This volume contains revised versions of the papers selected for the fourth volume of the Online Handbook of Argumentation for AI (OHAAI). Previously, formal theories of argument and argument interaction have been proposed and studied, and this has led to the more recent study of computational models of argument. Argumentation, as a field within artificial intelligence (AI), is highly relevant for researchers interested in symbolic representations of knowledge and defeasible reasoning. 
The purpose of this handbook is to provide an open access and curated anthology for the argumentation research community. OHAAI is designed to serve as a research hub to keep track of the latest and upcoming PhD-driven research on the theory and application of argumentation in all areas related to AI. The handbook’s goals are to:

\begin{enumerate}
    \item Encourage collaboration and knowledge discovery between members of the argumentation community.
    \item Provide a platform for PhD students to have their work published in a citable peer-reviewed venue.
    \item Present an indication of the scope and quality of PhD research involving argumentation for AI.
\end{enumerate}

The papers in this volume are those selected for inclusion in OHAAI Vol.\,4 following a single-blind peer-review process. The volume thus presents a strong representation of the contemporary state of the art research of argumentation in AI that has been strictly undertaken during PhD studies. Papers in this volume are listed alphabetically by author. We hope that you will enjoy reading this handbook.
\begin{flushright}
\noindent\begin{tabular}{r}
\makebox[1.3in]{}\\
\textit{The Editors}
\\Elfia Bezou-Vrakatseli, 
\\Federico Castagna,
\\Isabelle Kuhlmann, 
\\Jack Mumford, 
\\Daphne Odekerken, 
\\Stefan Sarkadi, 
\\Maddie Waller, 
\\Andreas Xydis
\\
\textbf{December 2023}
\end{tabular}
\end{flushright}

\pagenumbering{gobble}

\begin{center}
    \textbf{{\huge Acknowledgements}}
\end{center}

\hfill

\noindent
We thank the senior researchers in the area of Argumentation and Artificial Intelligence for their efforts in spreading the word about the OHAAI project with early-career researchers.

\hfill

\noindent
We are especially thankful to the COMMA organisers for collaborating with us since 2022.

\hfill

\noindent
We are also grateful to ArXiv for their willingness to publish this handbook.

\hfill

\noindent
Our sincere gratitude to Costanza Hardouin for her fantastic work in designing the OHAAI logo.

\hfill

\noindent
We owe many thanks to Sanjay Modgil for helping to form the motivation for the handbook, and to Elizabeth Black and Simon Parsons for their advice and guidance that enabled the OHAAI project to come to fruition.

\hfill

\noindent
The success of the OHAAI project depends upon the quality feedback provided by our reviewers. We have been fortunate in securing a diligent and thoughtful program committee that produced reviews of a reliably high standard. Our thanks go to our PC: 

\noindent
Lars Bengel, Lydia Bl{\"u}mel, Andreas Br{\"a}nnstr{\"o}m, Théo Duchatelle, Alex Jackson, Timotheus Kampik, Avinash Kori, Stipe Pand\v zi\' c, Guilherme Paulino-Passos, Christos Rodosthenous, Fabrizio Russo, Robin Schaefer, Antonio Yuste-Ginel, and Heng Zheng.

\hfill

\noindent
Special thanks must go to the contributing authors: 

\noindent
Lars Bengel, Elfia Bezou-Vrakatseli, Lydia Bl{\"u}mel, Giulia D’Agostino, Daphne Odekerken, Fabrizio Russo, Madeleine Waller. Thank you for making the world of argumentation greater! 


\newgeometry{margin = 0.9in}

\pagenumbering{arabic}

\tableofcontents 
\thispagestyle{empty}
\clearpage

\pagestyle{fancy}
\addtocontents{toc}{\protect\renewcommand{\protect\cftchapleader}
     {\protect\cftdotfill{\cftsecdotsep}}}
\addtocontents{toc}{\setlength{\protect\cftchapnumwidth}{0pt}}

\refstepcounter{chapter}\label{0}
\addcontentsline{toc}{chapter}{Editors' Note \\ \textnormal{\textit{Elfia Bezou-Vrakatseli, Federico Castagna, Isabelle Kuhlmann, Jack Mumford, Stefan Sarkadi, Daphne Odekerken, Maddie Waller, Andreas Xydis}}}
\includepdf[pages=-,pagecommand={\thispagestyle{plain}}]{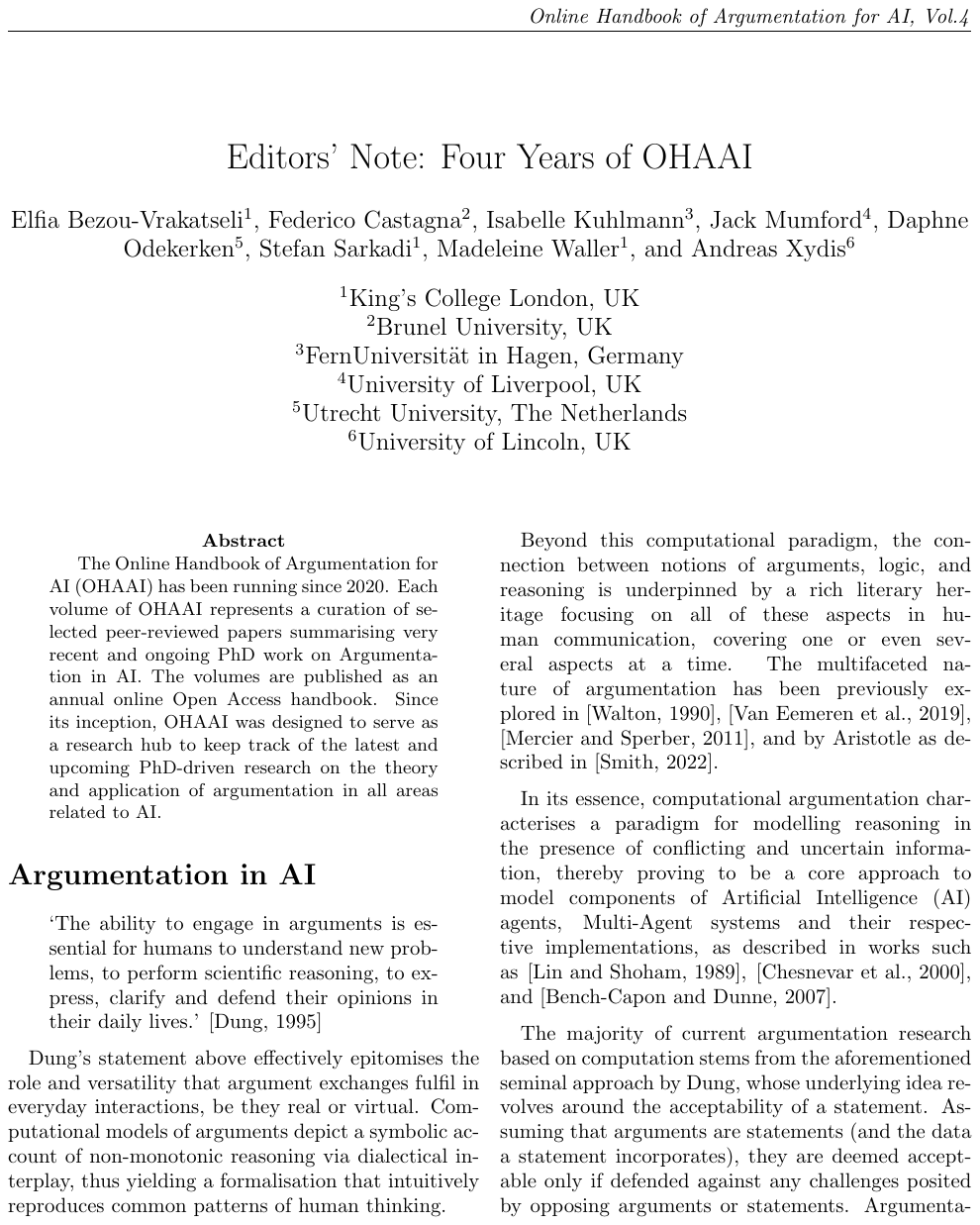}

\refstepcounter{chapter}\label{1}
\addcontentsline{toc}{chapter}{Towards Generalising Serialisability to other Argumentation Formalisms \\ \textnormal{\textit{Lars Bengel}}}
\includepdf[pages=-,pagecommand={\thispagestyle{plain}}]{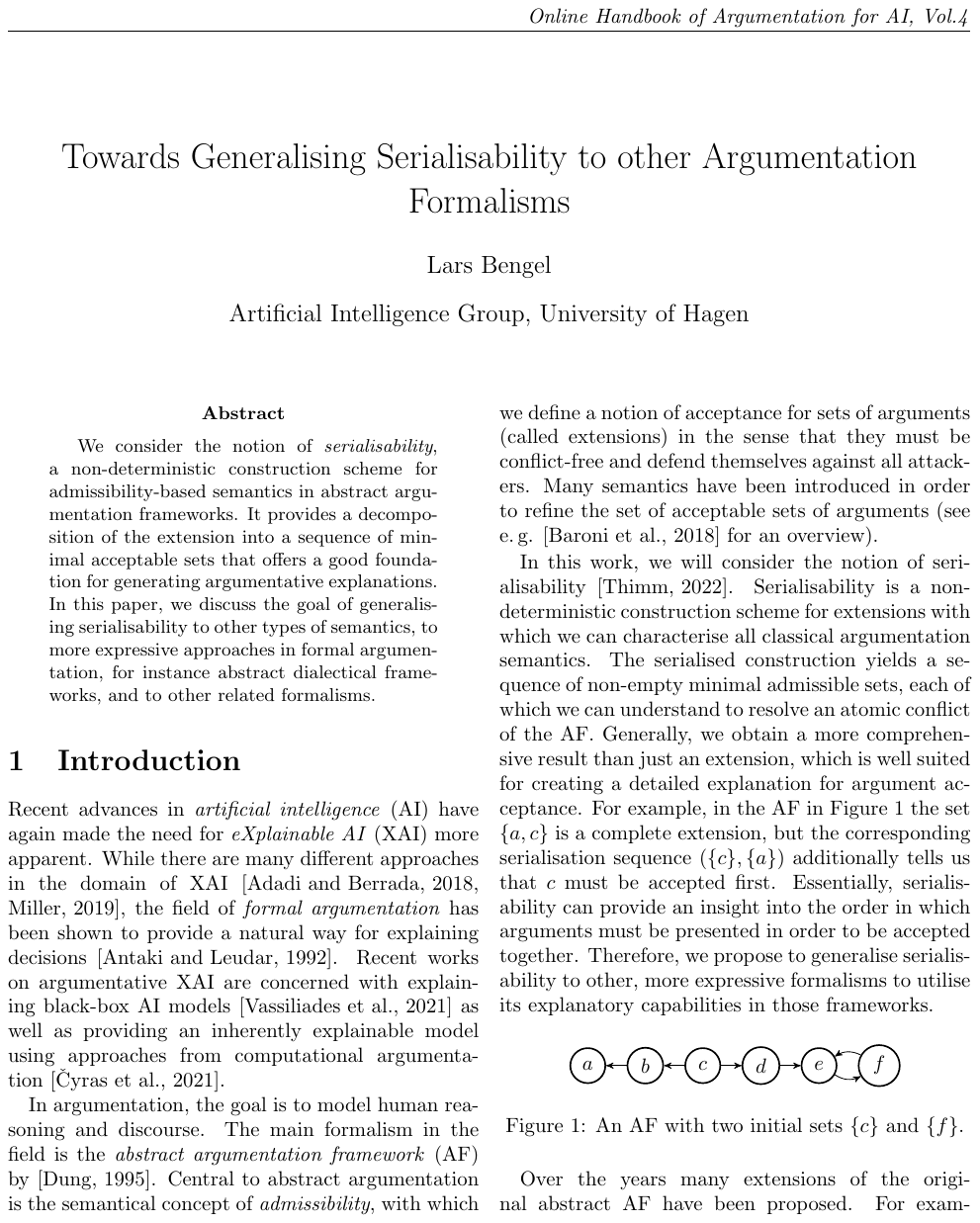}

\refstepcounter{chapter}\label{2}
\addcontentsline{toc}{chapter}{Evaluation of LLM Reasoning via Argument Schemes \\ \textnormal{\textit{Elfia Bezou-Vrakatseli}}}
\includepdf[pages=-,pagecommand={\thispagestyle{plain}}]{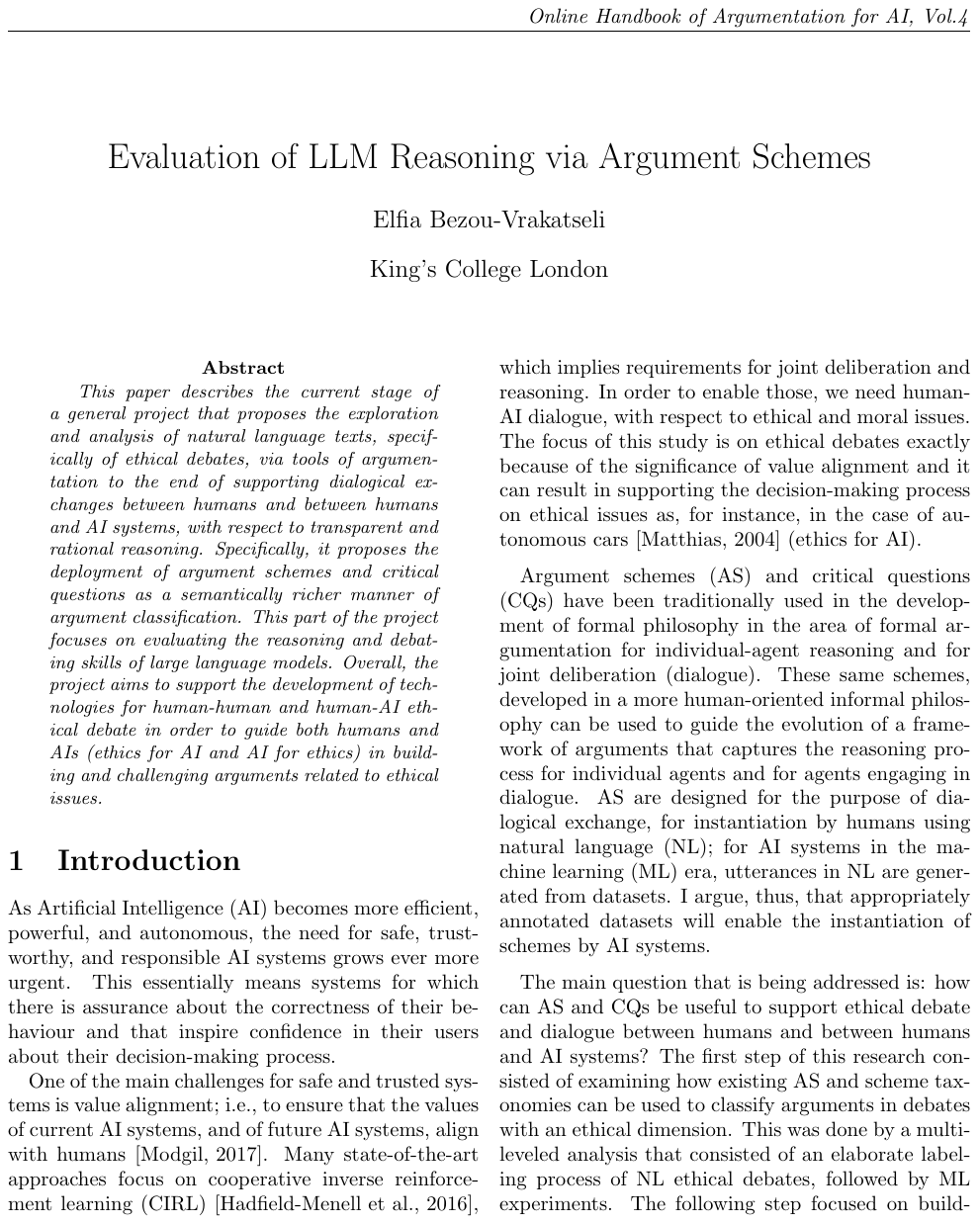}

\refstepcounter{chapter}\label{3}
\addcontentsline{toc}{chapter}{Refute-Based Representations of Recently Introduced Abstract Argumentation Semantics \\ \textnormal{\textit{Lydia Blümel}}}
\includepdf[pages=-,pagecommand={\thispagestyle{plain}}]{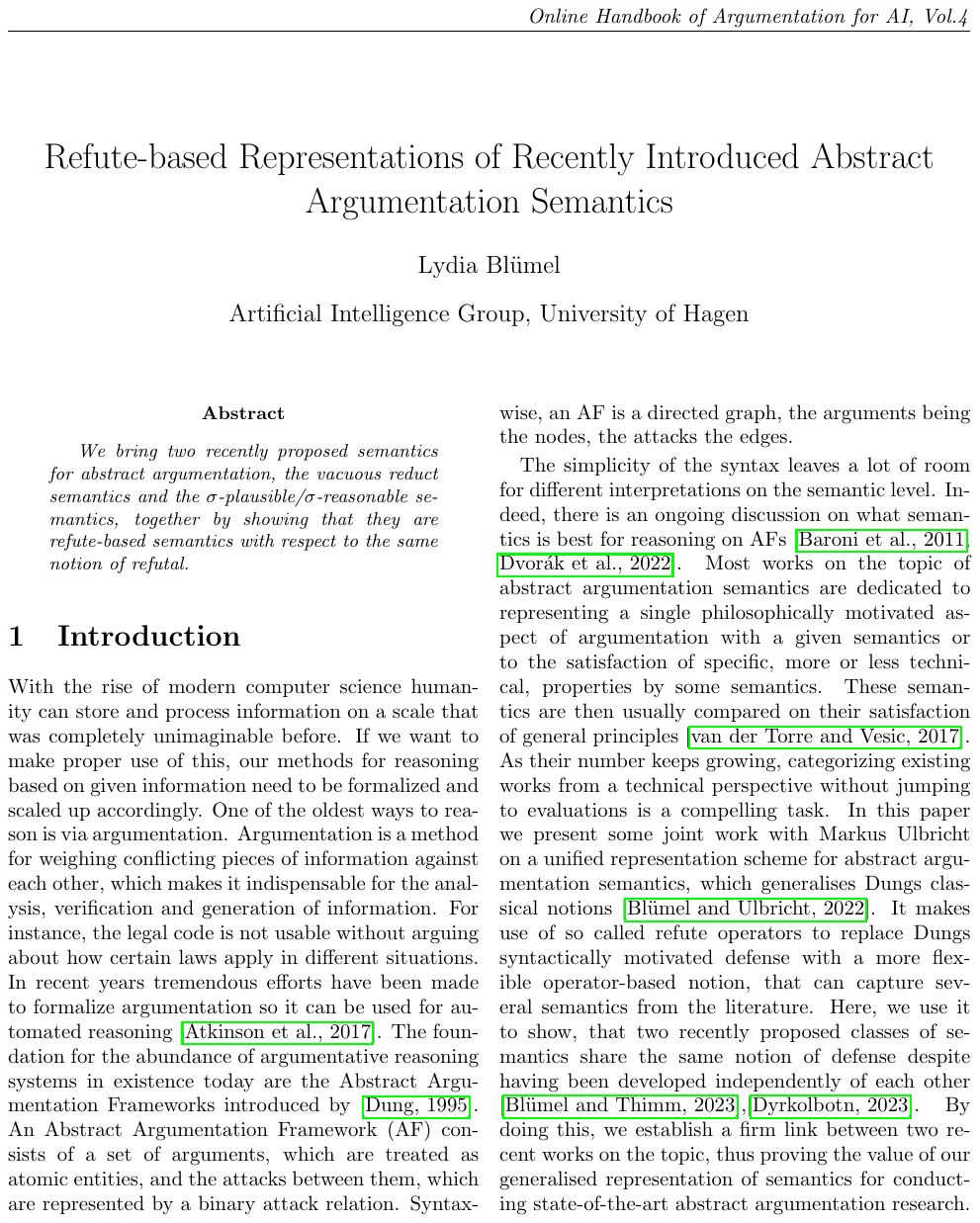}

\refstepcounter{chapter}\label{4}
\addcontentsline{toc}{chapter}{Opposition Without Argumentation \\ \textnormal{\textit{Giulia D'Agostino}}}
\includepdf[pages=-,pagecommand={\thispagestyle{plain}}]{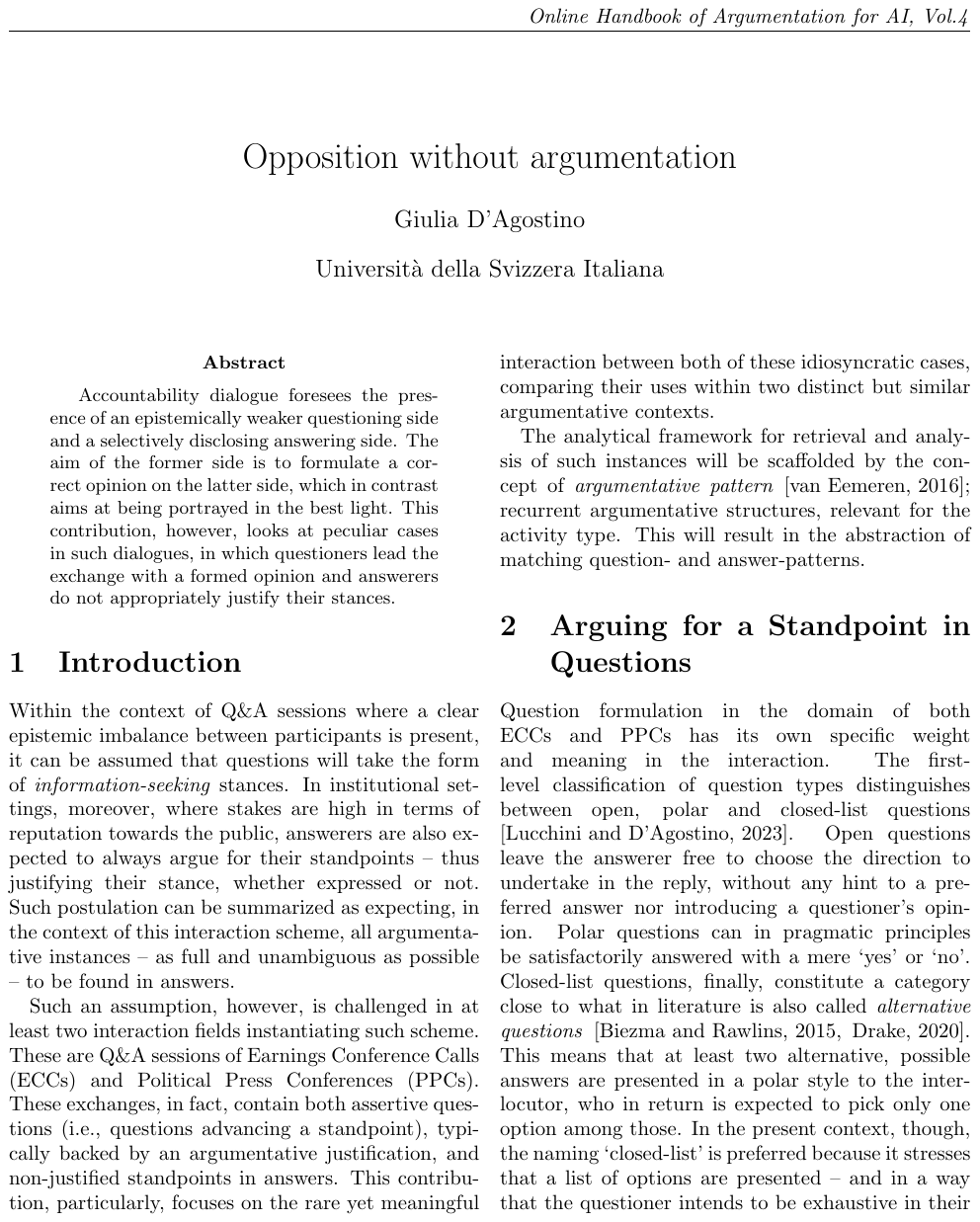}

\refstepcounter{chapter}\label{5}
\addcontentsline{toc}{chapter}{Argumentative Reasoning with Incomplete Information in Law Enforcement \\ \textnormal{\textit{Daphne Odekerken}}}
\includepdf[pages=-,pagecommand={\thispagestyle{plain}}]{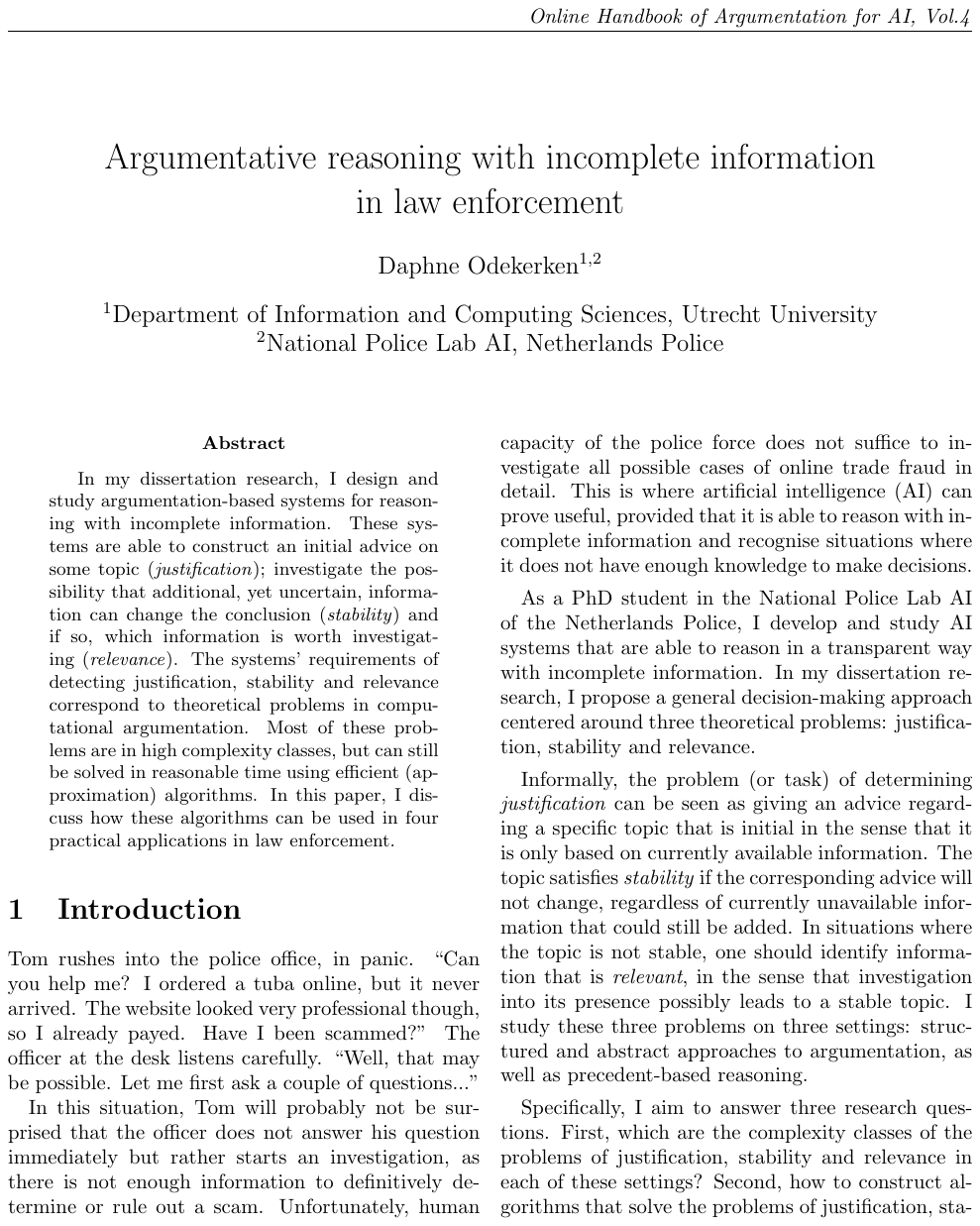}

\refstepcounter{chapter}\label{6}
\addcontentsline{toc}{chapter}{Argumentative Causal Discovery and Explanations \\ \textnormal{\textit{Fabrizio Russo}}}
\includepdf[pages=-,pagecommand={\thispagestyle{plain}}]{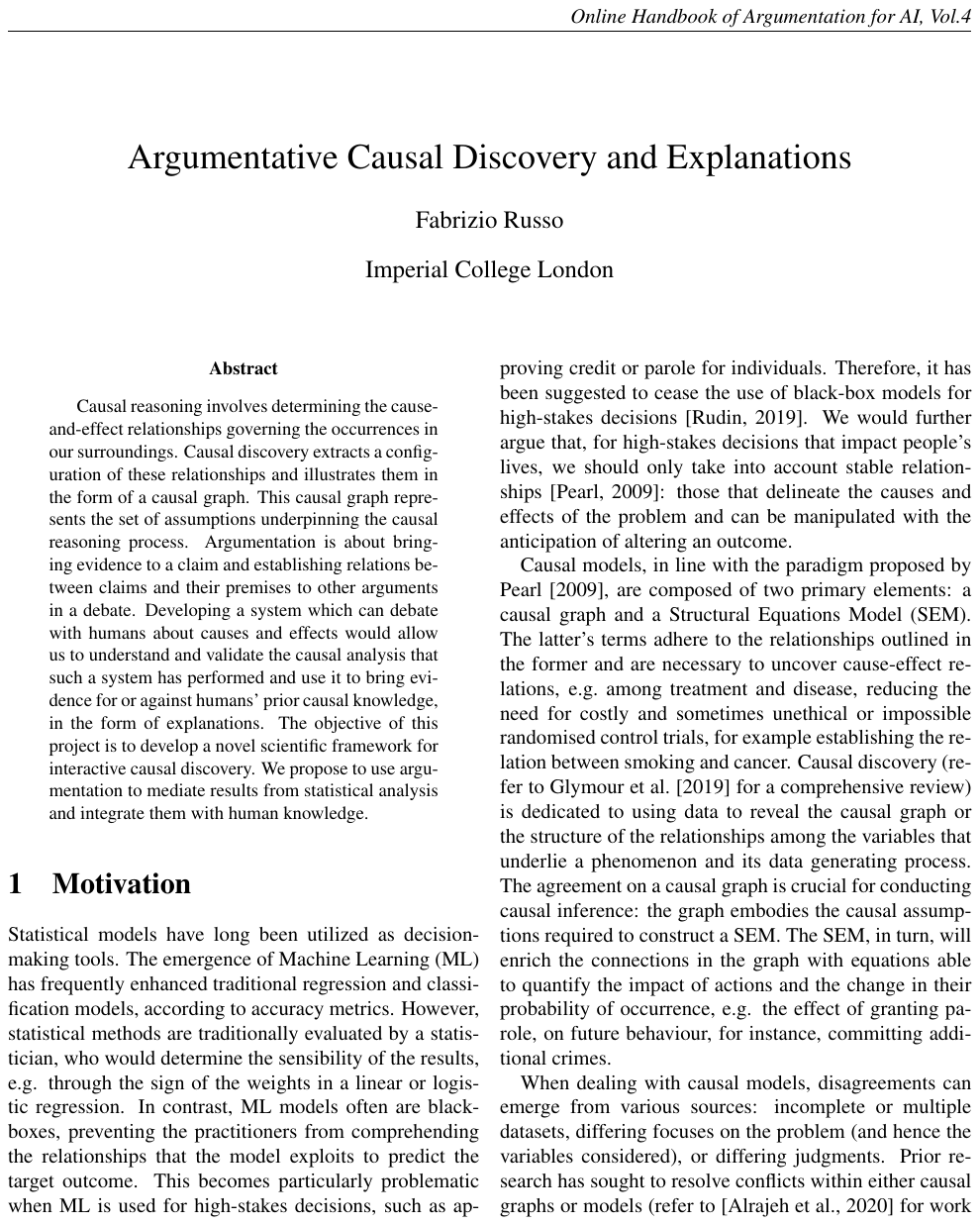}

\refstepcounter{chapter}\label{7}
\addcontentsline{toc}{chapter}{An Argumentation-Based Approach to Bias Detection in Automated Decision-Making Systems \\ \textnormal{\textit{Madeleine Waller}}}
\includepdf[pages=-,pagecommand={\thispagestyle{plain}}]{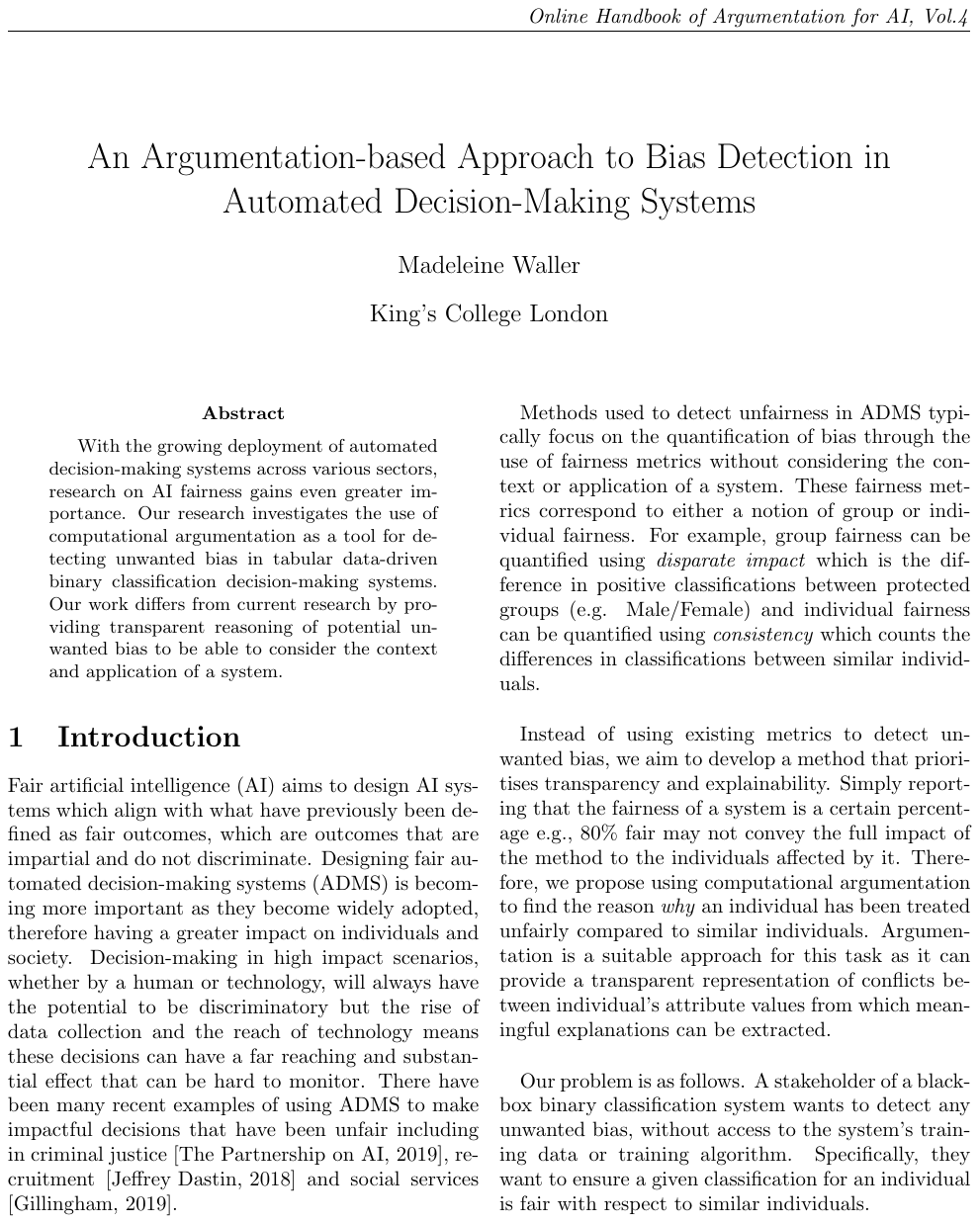}

\end{document}